\documentclass[conference]{IEEEtran}
\IEEEoverridecommandlockouts
\usepackage{cite}
\usepackage{amsmath,amssymb,amsfonts}
\usepackage{algorithmic}
\usepackage{graphicx}
\usepackage{textcomp}
\usepackage{xcolor}
\def\BibTeX{{\rm B\kern-.05em{\sc i\kern-.025em b}\kern-.08em
    T\kern-.1667em\lower.7ex\hbox{E}\kern-.125emX}}
\begin{document}

\title{Machine Learning Classification of Peaceful Countries: A Comparative Analysis and Dataset Optimization\\ \thanks{Funding provided by the Climate School at Columbia University for student researchers and from the Toyota Research Institute.}
}

\author{\IEEEauthorblockN{Kevin Lian}
\IEEEauthorblockA{\textit{Fu Foundation School of} \\  \textit{Engineering and Applied Science} \\
\textit{Columbia University}\\
New York, NY, USA \\
kl3451@columbia.edu}

\and
\IEEEauthorblockN{ Larry S. Liebovitch}
\IEEEauthorblockA{\textit{AC4 in the Climate School} \\
\textit{Columbia University}\\
New York, NY, USA \\
lsl2140@columbia.edu}

\and
\IEEEauthorblockN{Melissa Wild}
\IEEEauthorblockA{\textit{AC4 in the Climate School} \\
\textit{Columbia University}\\
New York, NY, USA \\
mm3484@columbia.edu}

\and
\IEEEauthorblockN{Harry West}
\IEEEauthorblockA{\textit{Indust.l Engr. \& Ops. Research} \\
\textit{Columbia University}\\
New York, NY, USA \\
hw2599@columbia.edu}

\and
\IEEEauthorblockN{Peter T. Coleman}
\IEEEauthorblockA{\textit{Teachers College,}  \\ \textit{AC4 in the Climate School} \\
\textit{Columbia University}\\
New York, NY, USA \\
coleman@exchange.tc.columbia.edu}

\and
\IEEEauthorblockN{Francine Chen}
\IEEEauthorblockA{\textit{Harmonious Communities,} \\ \textit{Human-Centered AI} \\
\textit{Toyota Research Institute}\\
Los Altos, CA, USA \\
francine.chen@tri.global}
\and
\IEEEauthorblockN{Everlyne Kimani}
\IEEEauthorblockA{\textit{Harmonious Communities,} \\ \textit{Human-Centered AI} \\
\textit{Toyota Research Institute}\\
Los Altos, CA, USA \\
everlyne.kimani@tri.global}
\and
\IEEEauthorblockN{Kate Sieck}
\IEEEauthorblockA{\textit{Harmonious Communities,} \\ \textit{Human-Centered AI} \\
\textit{Toyota Research Institute}\\
Los Altos, CA, USA \\
kate.sieck@tri.global}

}

\maketitle

\begin{abstract}
This paper presents a machine learning approach to classify countries as peaceful or non-peaceful using linguistic patterns extracted from global media articles. We employ vector embeddings and cosine similarity to develop a supervised classification model that effectively identifies peaceful countries. Additionally, we explore the impact of dataset size on model performance, investigating how shrinking the dataset influences classification accuracy. Our results highlight the challenges and opportunities associated with using large-scale text data for peace studies.
\end{abstract}

\section{Introduction}

Studies using social science methods and mathematical modeling are trying to understand the differences between peaceful and non-peaceful cultures \cite{ampsych},\cite{fry},\cite{lsl}. In the study of global peace, identifying and classifying peaceful countries through media representations is an essential task. Traditionally, research in this area has relied on word frequency analyses or content analyses limited by scale. However, with advancements in artificial intelligence, there is potential to perform more comprehensive and nuanced analyses using machine learning models. This paper aims to develop a machine learning classification model that can accurately identify peaceful and non-peaceful countries based on the linguistic patterns in media articles. 
The data used in this study was drawn from the News On the Web (NOW) dataset, which contains 700,000 articles comprising 58 million words. The mean number of articles per country across the 18 countries included in the analysis was 40,199, with a standard deviation of 24,214 articles. Similarly, the mean number of words per country was 3,212,191, with a standard deviation of 2,083,456 words. As described by Liebovitch et al. \cite{plosone}
\begin{quote} ``the dataset includes a wide range of topics such as accidents, business, crime, education, the arts, government, healthcare, law, literature, medicine, politics, real estate, religion, sports, war, as well as book, music, and movie reviews. Some examples of media sources included in the dataset are AlterNet, Business Insider, Chicago Tribune, USA TODAY, Jerusalem Post, and Vulture, among many others.''
\end{quote}

\subsection{Background}

Previous studies have demonstrated the importance of media representations in shaping public perceptions of peace. However, these studies often focused on limited datasets and utilized basic analytical methods, such as word frequency counts. Our study seeks to advance this research by leveraging vector embeddings to capture the semantic meaning of text, allowing for more sophisticated classification and analysis.

\subsection{Objectives and Contributions}

The primary contributions of this paper are twofold:

\begin{enumerate}
    \item \textbf{Development of a Supervised Classification Model:} We develop a model that classifies media articles as originating from peaceful or non-peaceful contexts using vector embeddings and cosine similarity. This model demonstrates high accuracy and provides insights into the linguistic patterns that are predictive of a country's peace level.
   
    \item \textbf{Dataset Optimization:} We explore the impact of dataset size on model performance, evaluating how reducing the number of articles used in training affects classification accuracy. This analysis is crucial for understanding the trade-offs between data volume and model performance in large-scale text analysis.
\end{enumerate}

\section{Methods}

This study employs a systematic approach to classify and analyze media articles using advanced machine learning techniques. The methodology is structured into three main components: embedding the articles, performing machine learning classification through cosine similarity, and exploring the effects of dataset size on model performance. This section provides a detailed explanation of each component.

\subsection{Embedding the Articles}

The embedding process transforms textual data into high-dimensional vectors, which are essential for similarity computations and classification. This process is carried out using Python scripts and involves several key steps:

\subsubsection{Data Preprocessing}

Before generating embeddings, we considered whether to preprocess the raw text data to ensure consistency and relevance in the embeddings. Traditional preprocessing steps include:

\begin{itemize}
    \item \textbf{Tokenization}: Splitting the text into individual tokens or words.
    \item \textbf{Stop Words Removal}: Removing common, uninformative words that do not significantly contribute to the semantic meaning.
    \item \textbf{Punctuation Removal}: Eliminating punctuation marks to focus solely on the textual content.
    \item \textbf{Stemming/Lemmatization}: Reducing words to their base or root forms to standardize different inflections of the same word.
\end{itemize}

After careful consideration, we chose to work with raw, unfiltered text. This decision was based on the understanding that large language models (LLMs), trained on extensive corpora of natural language, are inherently equipped to handle unprocessed text. This approach leverages the LLMs' capability to understand and contextualize natural language, potentially enhancing the quality and accuracy of the embeddings.

\subsubsection{Embedding Generation}

Each article is tagged with its country of origin and the peace level associated with that country. As determined from 5 independent peace indices described in Liebovitch et al. \cite{plosone}. The tagged text is then passed through a pre-trained embedding function, specifically the OpenAI text-embedding-3-small model. This function generates a 1536-dimensional vector for each article, capturing the semantic meaning of the text. These vectors are crucial for downstream tasks, such as similarity computations and classifications.

\subsubsection{Storage}

The generated embeddings are stored in a ChromaDB database, which facilitates efficient retrieval and comparison during the classification phase. This storage solution is essential for handling large volumes of data, ensuring quick access to embeddings when performing classification tasks.

\subsection{Machine Learning Classification}

The classification process is performed using a machine learning approach based on cosine similarity between the generated vectors. The steps involved are as follows:

\subsubsection{Query Embedding}

To classify a new article, the text is first embedded using the same pre-trained embedding function used for the training articles. This ensures that the query and training embeddings are situated within the same vector space, allowing for accurate similarity comparisons.

\subsubsection{Cosine Similarity Calculation}

Cosine similarity is calculated between the query embedding and each article embedding stored in the database. Cosine similarity measures how similar two vectors are, independent of their magnitude, and is defined as the cosine of the angle between two non-zero vectors. The similarity values range from -1 (opposite) to 0 (uncorrelated) to 1 (identical).

\subsubsection{Article Classification}

The article is classified based on its cosine similarity scores with the articles in the database. If the query article is more similar to articles tagged as coming from peaceful countries, it is classified as coming from a peaceful country, and vice versa. The classification model is evaluated using leave-out-one-cross-validation, where one country is excluded from the training set and used as the test set. This ensures that the model's performance is not biased towards any single country.

\subsubsection{Country Classification}

To maintain fairness in the comparison, the database contains an equal number of articles from each country. The number is determined by the country with the fewest available articles (Tanzania with 6000 articles). The overall peace level of a country is predicted by calculating the percentage of articles classified as peaceful out of the total number of articles for that country. A peace percentage over a set threshold indicates that the country is classified as peaceful, while a percentage the threshold indicates a non-peaceful classification.

\subsection{Impact of Dataset Size on Classification Performance}

Given the large scale of the dataset, it is important to understand how reducing the dataset size affects the classification model's performance. This section details the methodology for analyzing the impact of dataset size:

\subsubsection{Data Sampling}

To assess the effect of dataset size, we systematically reduce the number of articles used in the training set. The dataset is incrementally reduced by factors of two, starting from the full dataset down to the smallest possible subset that maintains at least one article per country. For each reduced dataset, we retrain the model and evaluate its performance using the same all-but-one testing strategy.

\subsubsection{Performance Metrics}

We evaluate the model's performance across different dataset sizes using standard classification metrics, including accuracy, precision, recall, and F1 score. These metrics are calculated for each iteration of dataset reduction, allowing us to observe trends and identify any significant changes in performance as the dataset size decreases.

\subsubsection{Visualization and Interpretation}

The results of the dataset size analysis are visualized using graphs that plot the performance metrics against the number of articles used in training (on a logarithmic scale). These visualizations help identify the point at which reducing the dataset size begins to significantly impact model performance, providing insights into the trade-offs between dataset size and classification accuracy.

\section{Results}

In this study, our classification model achieved an accuracy of 94\%. This high accuracy indicates the model's effectiveness in classifying articles as emanating from peaceful or non-peaceful contexts. The calculated peace percentages derived from the model's classifications correlated well with the Human Development Index (HDI), demonstrating the validity of our approach.

\subsection{Peace Percentage by Country}

Table 1 below shows the normalized peace percentages for each country included in our study:

\begin{table}[htbp]
\centering
\caption{Normalized Peace Percentages by Country}
\begin{tabular}{|l|l|}
\hline
\textbf{Country} & \textbf{Peace Percentage} \\
\hline
Australia & 76\% \\
Bangladesh & 18\% \\
Canada & 73\% \\
Ghana & 11\% \\
Hong Kong & 61\% \\
India & 33\% \\
Ireland & 66\% \\
Jamaica & 49\% \\
Kenya & 23\% \\
Malaysia & 57\% \\
New Zealand & 68\% \\
Nigeria & 11\% \\
Philippines & 30\% \\
Singapore & 48\% \\
Sri Lanka & 21\% \\
Tanzania & 13\% \\
United Kingdom & 70\% \\
United States & 65\% \\
\hline
\end{tabular}
\label{table:peace_percentages}
\end{table}

\subsection{Correlation with Human Development Index}

\begin{figure}[htbp]
    \centering
    \includegraphics[width=0.45\textwidth]{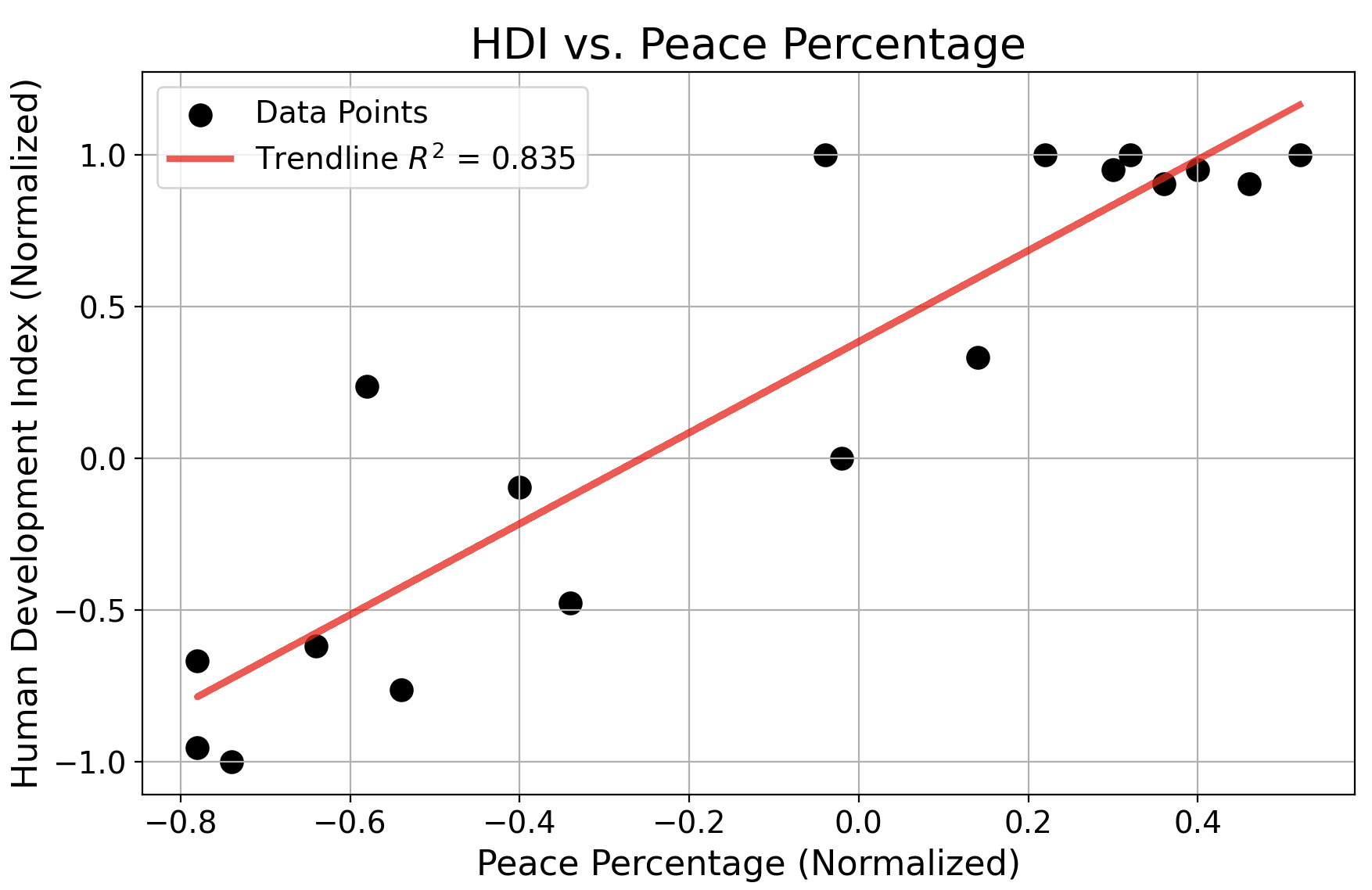}
    \caption{HDI vs. Peace Percentage}
    \label{fig:hdi_vs_peace}
\end{figure}

The scatter plot in Figure \ref{fig:hdi_vs_peace} illustrates the relationship between the normalized peace percentages and the Human Development Index (HDI) \cite{hdi}. The trendline shows a strong positive correlation (R² = 0.835), indicating that countries with higher peace percentages tend to have higher HDI scores.

This correlation suggests that our model's peace classification aligns well with established measures of human development, providing further evidence of the model's validity and potential utility for policymakers and researchers interested in the linguistic underpinnings of peace and development.

\subsection{Prediction of Peaceful and Non-Peaceful Countries}

Based on the calculated peace percentages, we classified each country as peaceful or non-peaceful. Countries with a peace percentage over 50\% are classified as peaceful, and those with a peace percentage under 50\% are classified as non-peaceful. The results are compared with our initial assumptions in Table \ref{table:peace_comparison}. These assumptions are obtained by averaging over 5 different indices that measure welfare of a country, as described in Liebovitch et al. \cite{plosone}. These indices include the Human Development Index, the Global Peace Index, the Positive Peace Index, the World Happiness Index, and the Fragile States Index.

\begin{table}[htbp]
\centering
\caption{Comparison of Predicted Peacefulness with Initial Assumptions}
\begin{tabular}{|l|l|l|}
\hline
\textbf{Country} & \textbf{Predicted} & \textbf{Initial Assumption} \\
\hline
Australia & Peaceful & Peaceful \\
Bangladesh & Non-Peaceful & Non-Peaceful \\
Canada & Peaceful & Peaceful \\
Ghana & Non-Peaceful & Non-Peaceful \\
Hong Kong & Peaceful & Peaceful \\
India & Non-Peaceful & Non-Peaceful \\
Ireland & Peaceful & Peaceful \\
Jamaica & Non-Peaceful & Non-Peaceful \\
Kenya & Non-Peaceful & Non-Peaceful \\
Malaysia & Peaceful & Peaceful \\
New Zealand & Peaceful & Peaceful \\
Nigeria & Non-Peaceful & Non-Peaceful \\
Philippines & Non-Peaceful & Non-Peaceful \\
Singapore & Non-Peaceful & Peaceful \\
Sri Lanka & Non-Peaceful & Non-Peaceful \\
Tanzania & Non-Peaceful & Non-Peaceful \\
United Kingdom & Peaceful & Peaceful \\
United States & Peaceful & Peaceful \\
\hline
\end{tabular}
\label{table:peace_comparison}
\end{table}

The comparison indicates that the model's predictions align well with our initial assumptions for most countries, with a minor discrepancy. Singapore is predicted differently compared to our initial assumptions, suggesting areas for further investigation and potential refinement of our classification model.

\subsection{Performance with Smaller Dataset}

\begin{figure}[htbp]
    \centering
    \includegraphics[width=0.45\textwidth]{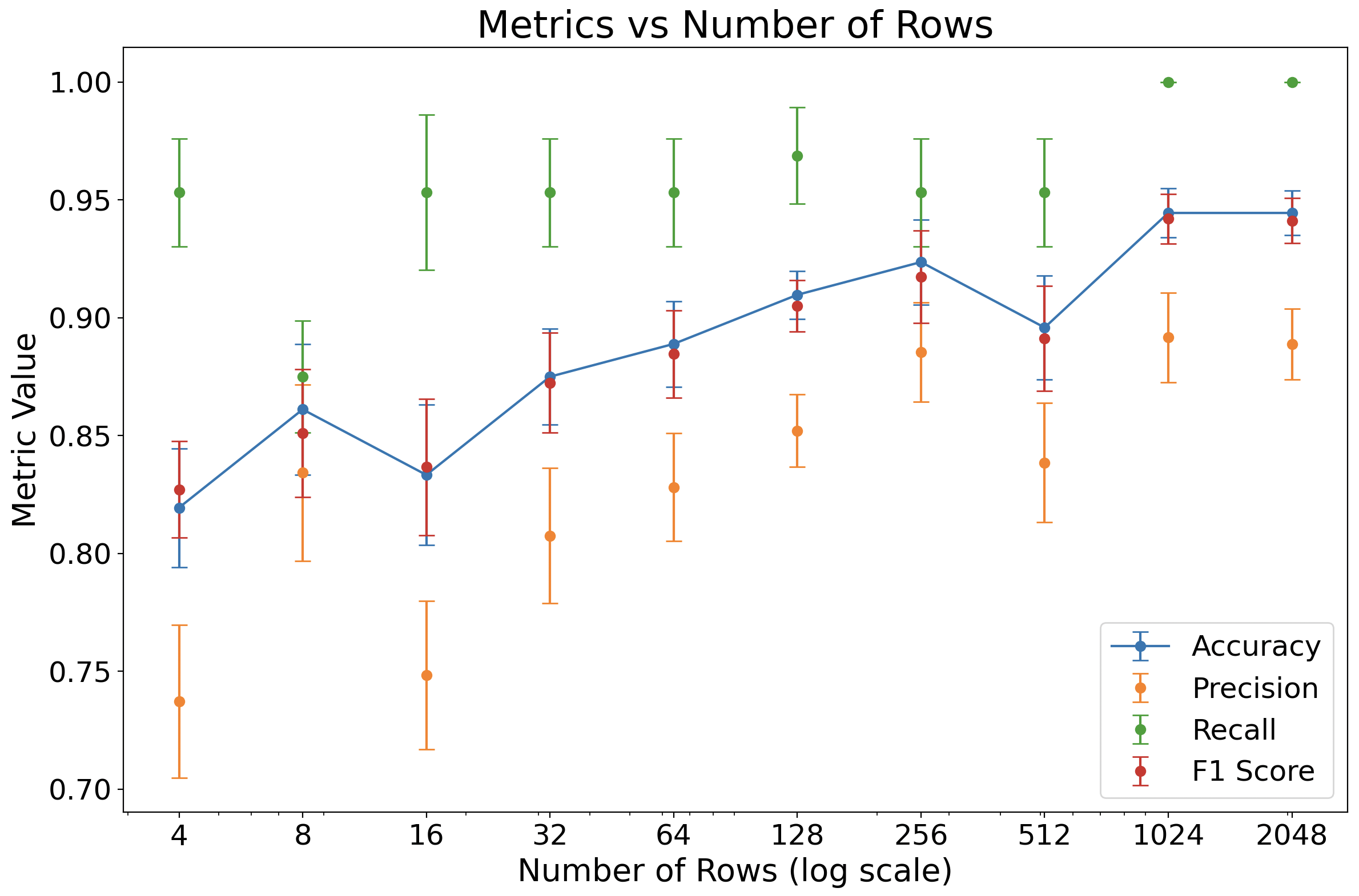}
    \caption{Metrics vs. the number of rows}
    \label{fig:metrics_vs_rows}
\end{figure}

In addition to our main analysis, we also explored the performance of our classification model with smaller datasets. Figure \ref{fig:metrics_vs_rows} illustrates the metrics (accuracy, precision, recall, and F1 score) against the number of rows (log scale) used in the training set.

This analysis shows that the model's performance varies with the size of the dataset. The training sets of data seem to be more biased towards peaceful countries, which is due to factors such as the training data containing more articles from peaceful countries and also articles from peaceful countries containing more general news topics instead of very localized topics, which could be more similar to other countries' articles.

\section{Discussion}

The results of our study provide valuable insights into the linguistic patterns associated with peaceful and non-peaceful nations. Our classification model demonstrated high accuracy in distinguishing between articles from peaceful and non-peaceful contexts, with a strong correlation between our calculated peace percentages and established measures of human development.

However, there are several potential limitations to our approach that should be acknowledged. One significant limitation is the dependency on the quality of the training data for the peace level classification. The effectiveness of our model relies heavily on the representativeness and accuracy of the training data. If the training data is biased or contains inaccuracies, the model's predictions may be skewed. For instance, our threshold percentage for classifying countries as peaceful versus non-peaceful is inherently an arbitrary number that has been adjusted to fit the training data. This means that the threshold may not be universally applicable or accurate across different datasets or contexts.

Moreover, the availability and volume of media articles from different countries can also impact the model's performance. Countries with more extensive media coverage may have more robust and diverse datasets, leading to more accurate classifications. Conversely, countries with limited media representation may result in less reliable predictions. This discrepancy underscores the importance of balanced and comprehensive data collection to ensure the model's validity.

\subsection{Comparison with Previous Work}

Our approach differs significantly from previous research, such as the study by Liebovitch et al. \cite{plosone} and others that relied on word frequency analyses to assess peace levels. By employing vector embeddings, we were able to capture the semantic meaning of the text, providing a more nuanced and comprehensive analysis. This method allows us to understand not just the presence of specific words, but the context in which they are used, leading to more accurate and insightful conclusions.

In comparison to studies that used word frequencies, our use of embeddings and advanced machine learning models offers a deeper understanding of linguistic patterns. This approach enables us to move beyond surface-level analysis and delve into the underlying themes and narratives that shape media representations of peace.

\subsection{Future Directions}

The promising results of this study open several avenues for future research and practical applications. One potential direction is the development of a real-time dashboard that monitors and visualizes peace levels across different countries. Such an application could leverage the methodologies developed in this study to provide policymakers, researchers, and the public with up-to-date insights into global peace dynamics.

Additionally, further research could explore the integration of other data sources, such as social media, to enhance the robustness and scope of the analysis. Expanding the dataset to include more countries and languages could also provide a more comprehensive understanding of global peace representations.

Another area for future work is the refinement of the model to address biases in the training data. As noted, the training sets currently show a bias towards peaceful countries, which may influence the model's performance. Developing strategies to mitigate this bias, such as balanced sampling or advanced normalization techniques, could improve the accuracy and fairness of the model.

Overall, our findings underscore the potential of using advanced AI techniques to analyze linguistic representations of peace in global media. By providing a comprehensive understanding of the factors influencing peace, our research offers valuable insights for policymakers, educators, and media professionals in shaping narratives that promote peace. The methodological approach demonstrated in this study can serve as a model for future research in peace studies and related disciplines, encouraging the use of AI and natural language processing in social sciences.

\end{document}